\documentclass[sigconf, nonacm]{acmart}
\usepackage{gensymb}
\usepackage{siunitx}
\usepackage{url}
\usepackage{flafter}
\usepackage[compact]{titlesec}
\usepackage{fancyhdr}

\begin{document}

\title{FireSRnet: Geoscience-Driven Super-Resolution of Future Fire Risk from Climate Change}
\setlength{\footskip}{25pt}
\lfoot{\fontsize{7}{12}\selectfont Tackling Climate Change with Machine Learning workshop at NeurIPS 2020}

%%
%% The "author" command and its associated commands are used to define the authors and their affiliations.
\author{Tristan Ballard}
\affiliation{%
  \institution{Sust Global}
  %\streetaddress{P.O. Box 1212}
 % \city{San Francisco}
 % \state{US}
}
\email{tristan@sustglobal.com}

\author{Gopal Erinjippurath}
%\orcid{0000-0002-1825-0097}
\affiliation{%
  \institution{Sust Global}
 % \streetaddress{1 Th{\o}rv{\"a}ld Circle}
 % \city{Hekla}
 % \country{Iceland}
}
\email{gopal@sustglobal.com}

\begin{abstract}
 With fires becoming increasingly frequent and severe across the globe in recent years, understanding climate change’s role in fire behavior is critical for quantifying current and future fire risk. However, global climate models typically simulate fire behavior at spatial scales too coarse for local risk assessments. Therefore, we propose a novel approach towards super-resolution (SR) enhancement of fire risk exposure maps that incorporates not only 2000 to 2020 monthly satellite observations of active fires but also local information on land cover and temperature. Inspired by SR architectures, we propose an efficient deep learning model trained for SR on fire risk exposure maps. We evaluate this model on resolution enhancement and find it outperforms standard image interpolation techniques at both 4x and 8x enhancement while having comparable performance at 2x enhancement. We then demonstrate the generalizability of this SR model over northern California and New South Wales, Australia. We conclude with a discussion and application of our proposed model to climate model simulations of fire risk in 2040 and 2100, illustrating the potential for SR enhancement of fire risk maps from the latest state-of-the-art climate models.
\end{abstract}
\maketitle

\section{Introduction}

Fires have become increasingly frequent and severe in recent years with catastrophic impacts \cite{johnston2020unprecedented, goss2020climate}. For example, the 2019-2020 Australian megafires burned 24m acres of vegetation habitat \cite{ward2020impact} and caused \$1.4b US in air quality health impacts \cite{johnston2020unprecedented}. In California, 5 of the state's top 6 largest fires have occurred in 2020 alone, including the first ever gigafire in modern history, defined as an individual fire burning over 1m acres \cite{CalFire}. The 2020 California fires have already burned over 4m acres, damaged or destroyed 10,000 structures, and caused 31 fatalities at the time of writing \cite{CalFireGeneral}. Accordingly, to what extent climate change has already impacted these fires, and whether it will increase the magnitude and locations of fire risk moving forward, is becoming crucially relevant in a range of sectors around the globe \cite{sanderson2020fiery, goss2020climate, barbero2015climate}.

Climate models provide an effective set of tools for quantifying the impact of climate change on acute physical risks like fire exposure \cite{eyring2016overview}. These models help identify regions with high risk and quantify the benefits of carbon emission reductions \cite{lasslop2020future}. However, a major limitation of the climate models from the Coupled Model Intercomparison Project phase 6 (CMIP6) is that they have spatial resolutions ranging from 0.5\degree{} to 2.5\degree{}, complicating local or asset-level analytics \cite{eyring2016overview, hantson2016status}. Enhancing spatial resolution on climate model fire exposure maps is therefore critical to making the exposure maps indicative of asset-level risk for both historical and forward-looking time horizons. 

\begin{figure}[] % t = top
\includegraphics[width=6.7cm]{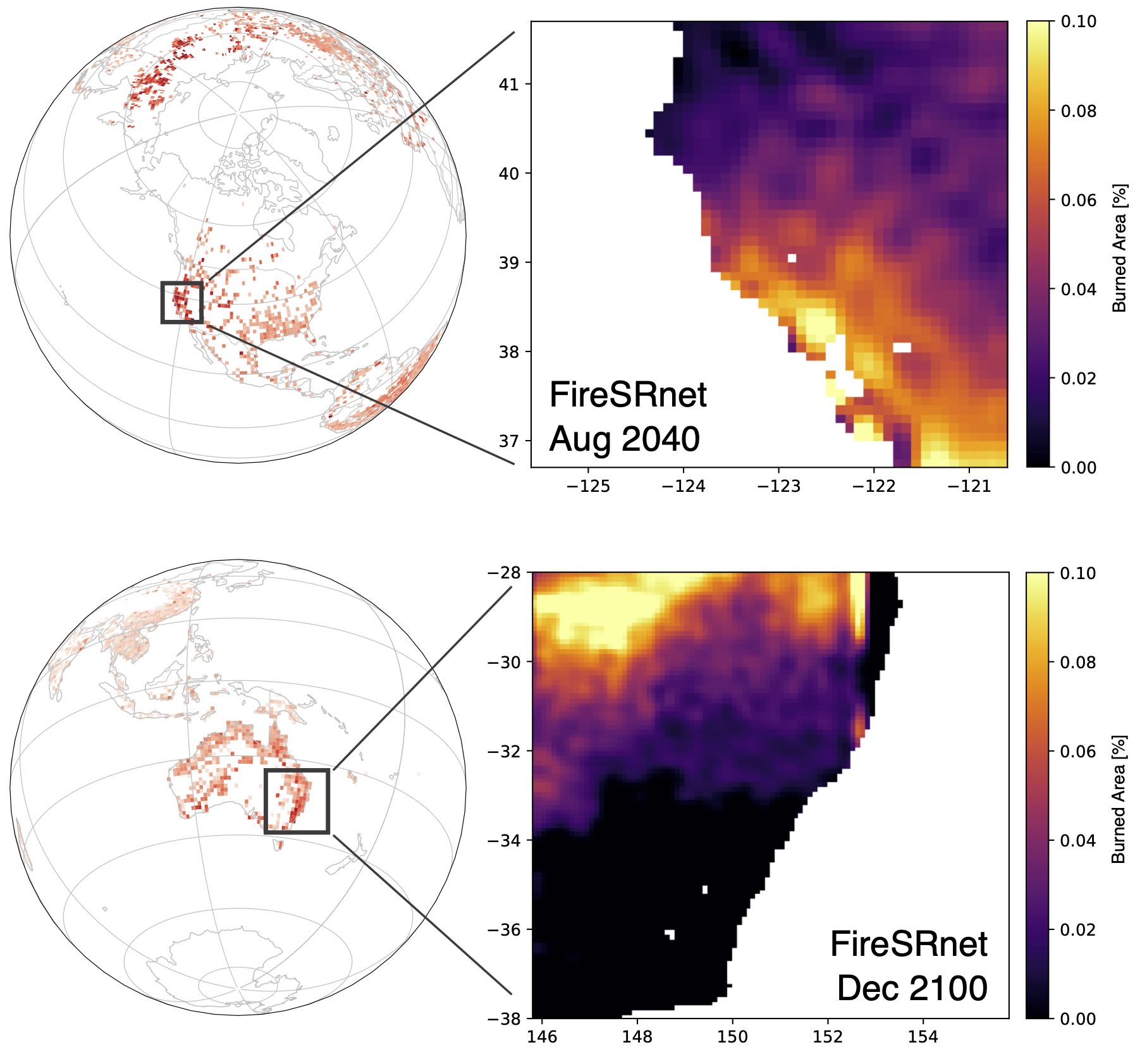}
\centering
\caption{Application of 4x (0.4\degree{} $\rightarrow$ 0.1\degree{}) SR FireSRnet to climate model simulations of fire exposure in two case study regions: northern California, US (top) and New South Wales, Australia (bottom).}
\label{fig:fig1}
\end{figure}

One promising approach for resolution enhancement is image super-resolution (SR), an area of model development gaining considerable attention in the AI and computer vision research community in recent years \cite{wang2018esrgan,  zhang2018residual, ledig2017photo}. While leading SR models can generate incredibly photo-realistic images, research efforts typically focus on real world visual imagery rather than geospatial datasets. However, since 2017 two studies have aimed to bring recent advances in SR modeling to the challenge of resolution enhancement of precipitation data, including from climate models \cite{wang2017deepsd, cheng2020reslap}. Those studies indicate an exciting potential for SR in the geosciences, but to date we are unaware of any such explorations beyond those with global precipitation data.

Here we propose FireSRnet, a novel SR architecture operating on a 3-channel geospatial dataset incorporating NASA satellite fire data, local temperature, and local land cover burnability. We compare FireSRnet performance at 2x, 4x, and 8x SR against a benchmark interpolation technique and validate model results with the recent fires in California and Australia. Finally, we showcase how FireSRnet can leverage CMIP6 climate model simulations of burned area and temperature to enable more precise forward-looking estimates of fire exposure (Fig. 1).

\section{Dataset creation}
Low-resolution (LR) geospatial data coded as input images to the SR model architecture (Fig. 2) are derived from high-resolution (HR) images having 3 channels corresponding to maps of fire counts, temperature deviation from mean conditions, and a burnable land index. We use monthly HR data at 0.1\degree{} ($\sim$\SI{11}km) resolution from March 2000 to August 2020 for both the continental US and Australia. Due to data quality concerns, we do not include March to July 2020 and March to May 2020 for the US and Australia, respectively. This results in a combined dataset of 240 images for the US and 243 images for Australia, each of size 256x584 pixels at HR.

\subsection{NASA satellite fire counts}
We use a monthly fire data product provided by the National Aeronautics and Space Administration (NASA) based on imagery from NASA's Terra and Aqua satellites \cite{NASAfire}. NASA’s global fire data product indicates the number of fires within a given 0.1\degree{} pixel each month and is available from March 2000 to present day. 

\subsection{Temperature deviation}
Because we expect temperature to be a key indicator of fire risk \cite{nolan2020causes}, we derived a monthly temperature deviation input channel using temperature data for the US and Australia. For the US, we use 4km resolution monthly temperature data provided by the PRISM group \cite{PRISM}. For Australia, we use 5km resolution monthly temperature data provided by the Australian Bureau of Meteorology \cite{jones2009high}. We expect temperature deviations, calculated by differencing each monthly grid cell relative to its 2000-2019 mean temperature, to be more informative than raw temperature since fires exhibit significant seasonality \cite{goss2020climate, nolan2020causes}. We then downsampled resulting temperature deviation maps to 0.1\degree{} using bilinear interpolation.

\subsection{Burnable land index}
We developed a burnable land index input channel based on satellite-derived land cover data with values ranging from 0 (low burnability) to 1 (high burnability). The 300m resolution land cover dataset created by the European Space Agency’s Land Cover Climate Change Initiative  assigns each pixel to one of 38 land cover classes based on 2015 satellite imagery \cite{li2018gross}. To make these land cover classes more scientifically relevant to the task of fire prediction, we binned each of the classes as either burnable (e.g. forest land cover types) or non-burnable (e.g. wetland). This binning was identical for the US and Australia with the exception of grassland and shrubland classes, which we classified as non-burnable and burnable for the US and Australia, respectively, due to Australia having large bushfires that are generally absent in the US for equivalent classes. We then downsampled the resulting binary map from 300m to 0.1\degree{} using bilinear interpolation, resulting in burnable land index values ranging from 0 to 1. We assume that the burnable land index is time-invariant across years evaluated in this study, in part because land use change at the spatial scales considered here has been relatively minimal during the study period \cite{fry2008completion, homer2015completion}. 

\begin{figure}
\includegraphics[width=\linewidth]{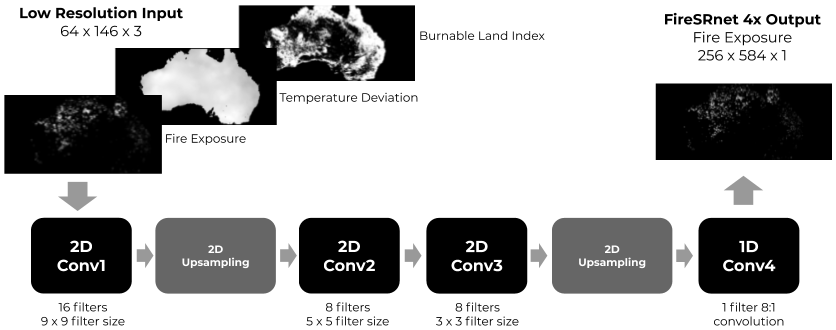}
\centering
\caption{FireSRnet model architecture for 4x SR, inputting 3 LR maps and outputting 1 SR map corresponding to fire exposure.}
\label{fig:fig2}
\end{figure}

\subsection{CMIP6 climate model simulations}
Climate models from leading international research centers are standardized through CMIP6 \cite{o2016scenario, eyring2016overview}. Here we use the CMIP6 Centre National de Recherches Météorologiques Earth System Model version 2.1 \cite{seferian2019evaluation} simulations of monthly temperature and burned area. To calculate the temperature deviation index, we use both historical temperature simulations spanning 2000 to 2015 and future simulations spanning 2016 to 2100 from the “Fossil-fueled Development” SSP5-RCP8.5 carbon emissions scenario \cite{riahi2017shared}, the scenario most consistent with current carbon emissions. The CMIP6 model employed here has a spatial resolution of 1.4\degree{}, which is regridded to the required LR input resolution.

\begin{figure*}[] % t = top, b = bottom
\centering
\includegraphics[width=\textwidth]{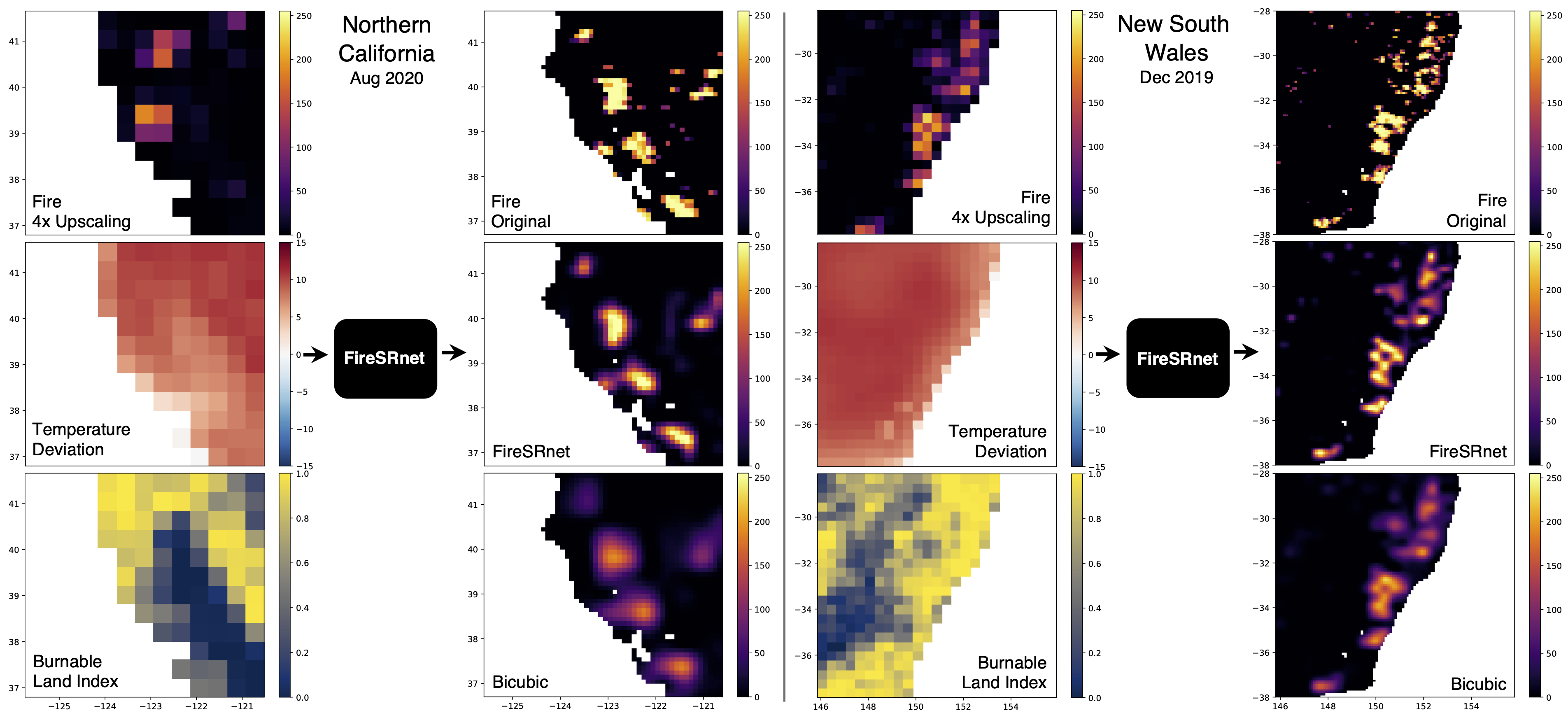}
\centering
\caption{Evaluation of 4x (0.4\degree{} $\rightarrow$ 0.1\degree{}) SR with out-of-sample fires in northern California (left) and New South Wales (right). FireSRnet outperforms bicubic interpolation in both fire magnitudes and spatial distribution. Input maps are NASA satellite fire counts [scaled 0-254], temperature deviation [\degree{}C], and burnable land index [ranging from low (0) to high (1) burnability]. }
\label{fig:fig3}
\end{figure*}

%%%%%%%%%%%%%%%%%%%%%%%%%%%%%%%%%%%%%%%%%%%%%%%%%%

\section{FireSRnet architecture}
Multiple approaches have been developed in the last few years for SR on color imagery \cite{wang2018esrgan,  zhang2018residual, ledig2017photo}. While the majority of such efforts focus on imagery of natural and man made objects, two recent efforts have applied SR algorithms to geospatial data, in particular to precipitation estimates at 2x, 4x, 5x and 8x SR enhancement \cite{wang2017deepsd, cheng2020reslap}. 

One important consideration for the design of our SR architecture is data availability. We use monthly data from 2000 to 2016 for training and model selection and 2017 to 2020 for out-of-sample quantitative and qualitative assessment of model performance. Due to the relatively small training dataset, we prioritized simple and efficient, albeit still high performing, deep learning architectures. 

Inspired by the approach described in \cite{dong2014learning}, we arrived at a fire super-resolution network (FireSRnet) deep learning architecture summarized in Figure 2 for 4x SR. The architecture interleaves 2D upsampling layers and 2D convolutional layers with variable size filters. We use progressively smaller 2D filters of size 9x9, 5x5 and 3x3 sizes in the first, second and third 2D convolutional layers, respectively with interleaved 2D 2x upsampling layers. The last stage uses a 1x1 convolution to combine learnings from multiple features maps to a single image at the target resolution. For the 2D convolutional layers, we use ReLU activation with the same padding, and for the upsampling layers we use bilinear interpolation. Owing to the small number of layers in this network, we have a total of 7.7K trainable parameters that can be readily trained from scratch. 

FireSRnet can be flexibly extended across different SR enhancement scales, and we have so far experimented with 2x, 4x and 8x SR (Figs. A1, A2). When going between SR scales, we use the same number of convolutional layers and trainable parameters. Lastly, since the goal of SR on risk exposure maps is to preserve the accuracy and fidelity of fire exposure, we use mean squared error throughout as our loss function rather than a perceptual loss function.   

Encouragingly, the layer 1 weights post-training result in a combination of spot detectors, offset spot detectors, spot eliminators, and sharpening filters, indicative of essential functions at the early stages of the network to transform the LR inputs and enable downstream feature maps that are indicative of discriminating features for fire detection (Fig. A3).

\section{Quantitative model evaluation}
For quantitative assessment of the performance of SR, we benchmark FireSRnet against bicubic interpolation by comparing the target HR image from which the LR input image was derived with the output SR or HR image calculated by FireSRnet or bicubic interpolation, respectively. 

We consider two aspects of model performance both relevant to fire exposure modeling. First, since fire exposure is a continuous value, we use R\textsuperscript{2} and RMSE to indicate how we track model prediction of the exposure magnitude as compared to the original HR exposure map. Second, we quantify the model's skill as a fire detector by converting both the HR target image and the predicted output images into binary fire maps. With these binary maps, we considering classification error metrics of precision, F1 score, and threat score \cite{cheng2020reslap}. We identify precision in particular as a key indicator of SR performance as it inversely correlates to false-positive detections. False-positive detections are the most common source of error when upsampling sparse LR exposure maps because upsampling diffuses exposure magnitude erroneously to neighborhood pixels.

\begin{table}[]
\centering
\begin{tabular}{@{}llllll@{}}
\toprule
            & RMSE              & R2               & Precision         & F1                & Threat Score      \\ \midrule
FireSRnet-2x & \textbf{0.0348} & \textbf{0.3810} & 0.9181          & 0.9567          & 0.9174          \\
Bicubic-2x   & 0.0351           & 0.3687          & \textbf{0.9284} & \textbf{0.9625} & \textbf{0.9281} \\
FireSRnet-4x & \textbf{0.0400} & \textbf{0.2434} & \textbf{0.9257}  & \textbf{0.9479} & \textbf{0.9015} \\
Bicubic-4x   & 0.0433         & 0.181           & 0.8747          & 0.9320          & 0.8735          \\
FireSRnet-8x & \textbf{0.0433} & \textbf{0.1218} & \textbf{0.8876} & \textbf{0.8998} & \textbf{0.8191} \\
Bicubic-8x   & 0.0473           & 0.104           & 0.8181          & 0.8971          & 0.8147          \\ 

\bottomrule
\end{tabular}
\caption{Performance of FireSRnet for 2x, 4x and 8x SR as compared to corresponding bicubic interpolation.}
\end{table}

\begin{figure*}[t] % t = top, b = bottom
\centering
\includegraphics[width=\textwidth]{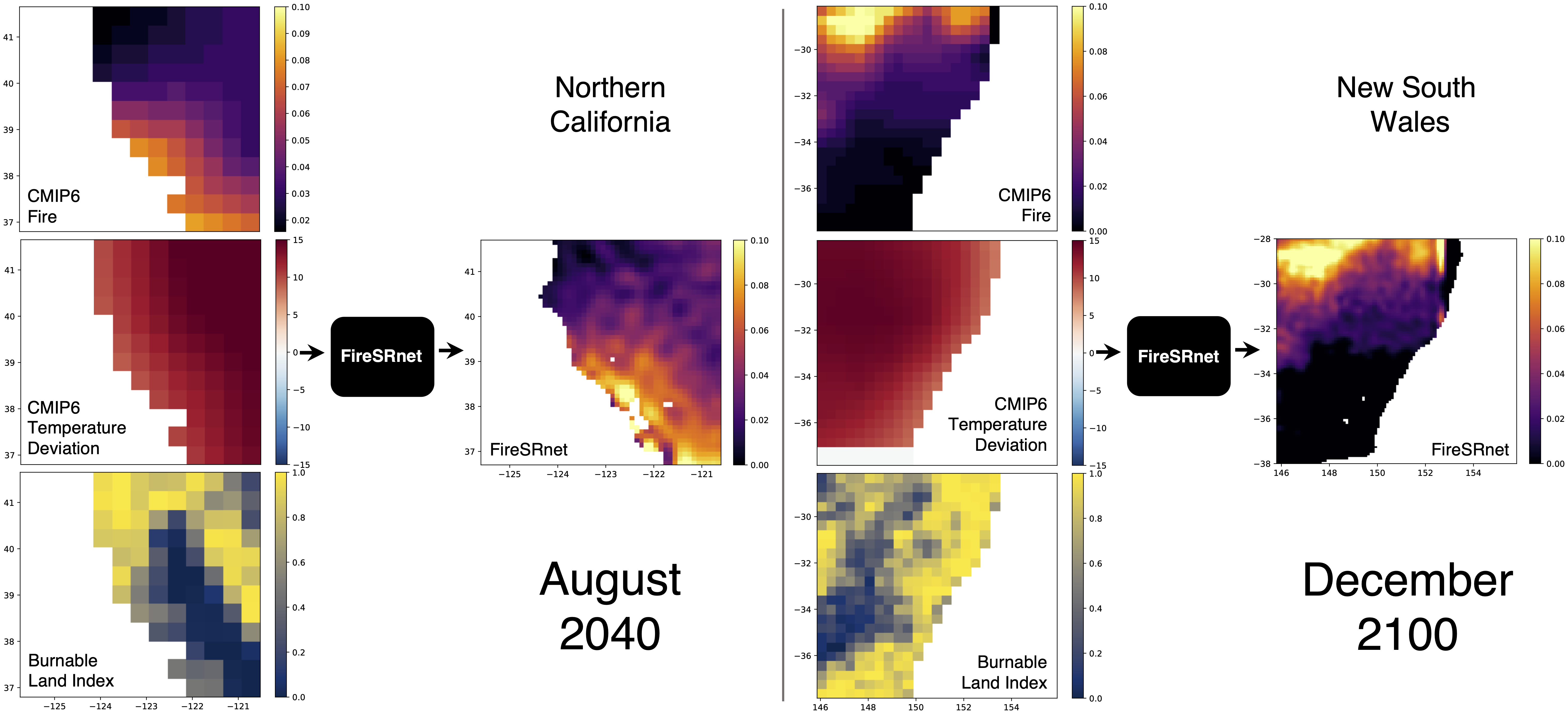}
\centering
\caption{Application of 4x (0.4\degree{} $\rightarrow$ 0.1\degree{}) SR to CMIP6 simulations of fire exposure in northern California (left) and New South Wales (right) for August 2040 and December 2100, respectively. Input image channels are CMIP6 burned area [\%], CMIP6 temperature deviation [\degree{}C], and a burnable land index [ranging from low (0) to high (1) burnability] and output is an single channel CMIP6 burned area SR image.}
\label{fig:fig4}
\end{figure*}

We find that FireSRnet outperforms bicubic interpolation for 4x and 8x enhancement on all metrics (Table 1). For 2x enhancement, FireSRnet outperforms bicubic interpolation for the metrics related to exposure magnitude, whereas bicubic interpolation outperforms FireSRnet on all three classification metrics (Table 1). This indicates that for 2x fire exposure enhancement, the benefits of FireSRnet over an interpolation alternative are less pronounced.

Next, we are interested in quantifying how generalizable FireSRnet is to different global regions and thus different ecosystems, climate conditions, and fire dynamics. To that end, we ran two additional experiments for region specific models: 1) Retrained on US data and validated out-of-sample on US data, and 2) Retrained  on Australia data and validated out-of-sample Australia data. Preliminary performance results (Table A1) between these two experiments and the full model provides evidence that the model could be generalized to regions beyond the US and Australia.

\section{Qualitative model evaluation}
We conduct case studies in northern California and New South Wales, visually comparing observed HR fire count maps with corresponding maps derived from FireSRnet and bicubic interpolation (Fig. 3). Because the case studies correspond to fire events in 2019 and 2020, neither were included in the training of FireSRnet. We focus here on the 4x SR maps, though the 2x and 8x SR maps indicate similar qualitative performance (not shown).

We find that FireSRnet outperforms bicubic interpolation in capturing both the magnitude and spatial distribution of the August 2020 California fires and the December 2019 New South Wales fires (Fig. 3). FireSRnet accurately captures the magnitudes of fire counts, whereas bicubic interpolation underestimates the magnitudes. This suggests that bicubic interpolation underestimates fire exposure, particularly in locations with the greatest fire activity. FireSRnet also appears to better capture the boundaries of large fires in both regions (Fig. 3), whereas bicubic interpolation results in overly smoothed fire boundaries. The better delineation of fire boundaries by FireSRnet is indicative of the deep learning model learning edge behavior between fire and no fire pixels (Fig. A3). 

Both FireSRnet and bicubic interpolation fail to capture fires with small spatial footprints (single 0.1\degree{} pixel size) since the signal from these small fires tends to be lost when creating the downsampled inputs (Fig. 3).

\section{Super-resolution of future fire exposure maps}
One of the most promising applications of the FireSRnet modeling framework (Fig. 2) is to enhance the resolution of fire exposure maps simulated by global climate models. Compared with enhancing CMIP6 maps through bicubic interpolation, FireSRnet incorporates additional geoscience information on land cover and future temperature when conducting the upsampling, suggesting it can more accurately upsample CMIP6 fire exposure maps. 

Here we display the potential of these SR fire exposure maps by applying the 4x SR FireSRnet, trained and validated on observed fire data, to CMIP6 simulations of future burned area and temperature in August 2040 and December 2100 for northern California and New South Wales, respectively (Fig. 4). For the burnable land index, we use the same 2015-derived input map for all years rather than modeling changes in land cover decades into the future.

Quantitative assessment and verification of the output SR maps is, of course, not possible for 2040 and 2100; nevertheless, the SR maps appear qualitatively reasonable. Interestingly, the northern California SR image suggests heightened risk in the Napa and Sonoma areas (38.2\degree, -122.5\degree) in 2040 (Fig. 4), an area experiencing fires as of October 2020. Meanwhile, the New South Wales SR image suggests mostly low fire exposure in 2100, particularly towards the south. 

While the observed HR fire count data employed to train FireSRnet is not strictly equivalent to the CMIP6 burned area data product, we argue that the pre-trained FireSRnet is suitable in this application. First, NASA designed the fire count data for a range of research applications, including validating global fire models \cite{NASAfire}. Second, qualitative assessment of the case study fire maps indicates that the NASA fire counts correspond well to burned area footprints (Fig 3), and qualitative assessment of the SR results (Fig. 4) further supports the generalization of FireSRnet to future CMIP6 fire simulation.

\section{Future directions}
We see multiple avenues for building on the FireSRnet modeling approach in the future. First, we intend to incorporate additional global regions including Siberia and the Amazon rainforest in our model training and assessments. Second, because FireSRnet can flexibly integrate additional input channels, we could include drought maps in the framework. Third, the rare event nature of the data suggests the potential of a zero-inflated, two-stage SR model \cite{diaz2019predicting}. Last, while we focus here on single image SR, we see opportunities for improved performance using multi-temporal SR, which has been applied to satellite imagery in the past \cite{deudon2020highres} but never, to our knowledge, to fire exposure maps.

\clearpage

%\begin{acks}
% This work was supported by the [...] Research Fund of [...] %(Number [...]). Additional funding was provided by [...] and %[...]. We also thank [...] for contributing [...].
%\end{acks}

%\clearpage

\bibliographystyle{unsrt}
%\bibliography{main}

\clearpage
\setcounter{table}{0}
\setcounter{figure}{0}
\renewcommand{\thetable}{A\arabic{table}}
\renewcommand\thefigure{A\arabic{figure}}    
\section{Appendix}

\begin{figure*}
\includegraphics[width=\textwidth]{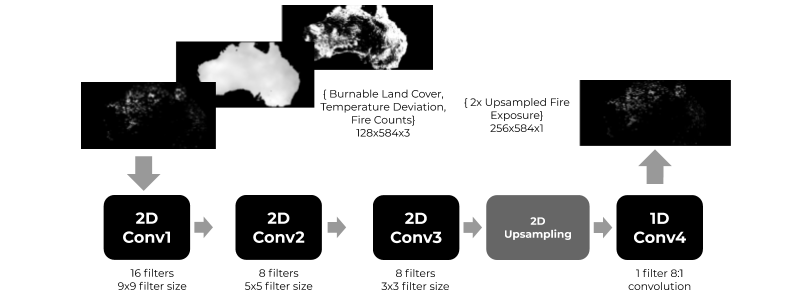}
%\centering
\caption{FireSRnet model architecture for 2x SR, inputting 3 LR maps and outputting 1 SR map corresponding to fire exposure.}
\label{fig:fig2}
\end{figure*}

\begin{figure*}
\includegraphics[width=\textwidth]{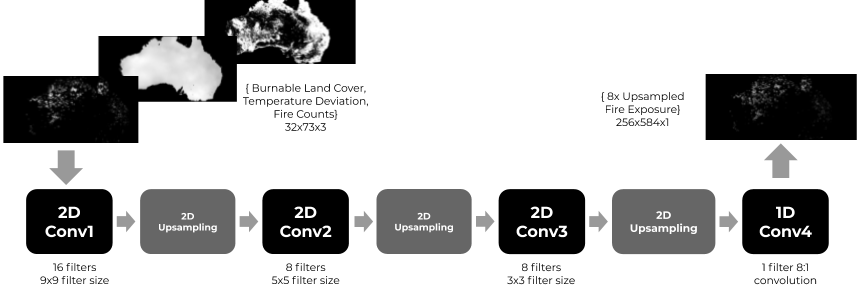}
%\centering
\caption{FireSRnet model architecture for 8x SR, inputting 3 LR maps and outputting 1 SR map corresponding to fire exposure.}
\label{fig:fig2}
\end{figure*}
\clearpage

\begin{figure*}
\includegraphics[width=\textwidth]{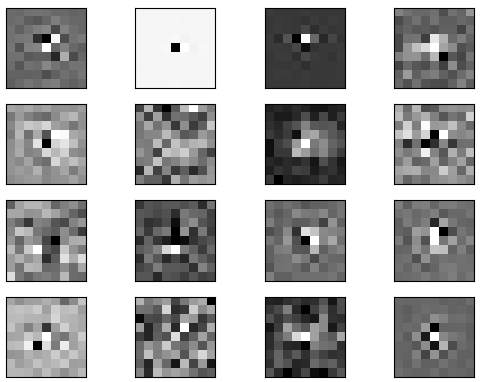}
%\centering
\caption{FireSRnet Layer 1 weights from the 16 2D conv1 filters.}
\label{fig:fig2}
\end{figure*}
\clearpage

\begin{table*}[t]
\centering
\begin{tabular}{@{}llllll@{}}
\toprule
Model/Metric          & RMSE     & R2       & Precision & F1       & Threat Score \\ \midrule
FireSRnet (US)    & 0.0390 & 0.2215 & 0.9157    & 0.9397 & 0.8870     \\
FireSRnet (AUS)    & 0.0423 & 0.1938 & 0.9307  & 0.9445  & 0.8951     \\ 
FireSRnet (US\_AUS) & 0.0398 & 0.2434 & 0.9257   & 0.9479 & 0.9015     \\ \bottomrule
\end{tabular}
\caption{Performance of FireSRnet for 4x SR on out of sample inference when trained on fire risk exposure maps from specific regions: US only, AUS only, US and AUS combined. The US\_AUS metrics are identical to those in the third row of Table 1.}
\end{table*}

\end{document}